%% file: main.tex
\documentclass{article}
\usepackage{spconf,amsmath,graphicx}
\usepackage{arabtex}
\usepackage{utf8}

\usepackage{booktabs}
\usepackage{amssymb}
\usepackage{amsfonts}
\usepackage{multicol}
\usepackage{multirow}
\usepackage[colorinlistoftodos]{todonotes}
\usepackage{hyperref}

\usepackage{arabtex}
\usepackage{utf8}

\usepackage{times}
\usepackage{url}
\usepackage{latexsym}
\usepackage{graphicx}
\usepackage{float}
\restylefloat{table}
\usepackage{enumitem}
\usepackage{color,soul}
\usepackage{verbatim}
\usepackage{arydshln}
\usepackage{textcomp}

\setcode{utf8}

\title{Speech Recognition Challenge in the Wild: Arabic MGB-3}
\makeatletter
\def\name#1{\gdef\@name{#1\\}}
\makeatother \name{Ahmed Ali$^{1,2}$, Stephan Vogel$^1$, Steve Renals$^2$}

\address{$^1$Qatar Computing Research Institute, HBKU, Doha, Qatar \\
$^2$Centre for Speech Technology Research,
University of Edinburgh, UK \\
  {\footnotesize mgb-admin@inf.ed.ac.uk www.mgb-challenge.org}
}
\begin{document}
%
\maketitle
\begin{abstract}
This paper describes the Arabic MGB-3 Challenge -- Arabic Speech Recognition in the Wild. Unlike last year's Arabic MGB-2 Challenge, for which the recognition task was based on more than 1,200 hours broadcast TV news recordings from Aljazeera Arabic TV programs, MGB-3 emphasises dialectal Arabic using a multi-genre collection of Egyptian YouTube videos. Seven genres were used for the data collection: comedy, cooking, family/kids, fashion, drama, sports, and science (TEDx). A total of 16 hours of videos, split evenly across the different genres, were divided into adaptation, development and evaluation data sets. The Arabic MGB-Challenge comprised two tasks: A) Speech transcription, evaluated on the MGB-3 test set, along with the 10 hour MGB-2 test set to report progress on the MGB-2 evaluation;  B) Arabic dialect identification, introduced this year in order to distinguish between four major Arabic dialects -- Egyptian, Levantine, North African, Gulf, as well as Modern Standard Arabic. Two hours of audio per dialect were released for development and a further two hours were used for evaluation. For dialect identification, both lexical features and i-vector bottleneck features were shared with participants in addition to the raw audio recordings.  Overall, thirteen teams submitted ten systems to the challenge.  We outline the approaches adopted in each system, and summarise the evaluation results.

\end{abstract}

\section{Introduction}
\input{intro}


\section{MGB-3 data}
\input{data}




\section{Baseline results}
\input{baseline}


\section{Submitted Systems and Results}
\input{participant}


\section{Conclusions}
The MGB-3 Arabic Challenge continued our efforts to evaluate speech recognition systems for diverse broadcast media, using fixed training sets. This year's challenge added a new task:  Arabic dialect identification (ADI) using both acoustic and linguistic information, where participants were asked to label each utterance with a dialect. The MGB-3 Arabic Challenge advanced the state-of-the-art with the best system achieving an average of 80\% overall accuracy across five classes using GANs and combining acoustic and lexical information. The speech-to-text transcription task substantially increased diversity by focusing on non-orthographic dialectal Arabic (DA), using a multi-genre YouTube collection across seven genres.  The Egyptian dialectal Arabic speech-to-text transcription task comprised only 16 hours of data split into three groups: adaptation, development, and testing, and participants needed to combine this data with the large MGB-2 training set.  Techniques resulting in reduced error rates included combining multiple reference transcripts using confusion networks for training, and the use of sub-word language models.  The most accurate result on the MGB-2  test set was 13.2\% WER (a 10\% relative improvement over the most accurate system in the 2016 MGB-2 Challenge), and the best result for the MGB-3 test data was 37.5\% average WER (AV-WER) and 29.3\% multi-reference WER (MR-WER). We plan to continue the challenge by adding more dialects and potentially collect more YouTube recording to explore transfer learning using a large pool of in-domain un-transcribed speech data.

\bigskip\noindent
\textbf{Acknowledgment:}
This work was partially supported by the H2020 project SUMMA, under grant agreement 688139.



\newpage
\bibliographystyle{IEEEbib}
\bibliography{strings,refs}

\end{document}

%% file: intro.tex
Following the previous MGB challenges \cite{bell2015mgb,ali2016mgb}, the MGB-3 Arabic Challenge was a controlled evaluation of speech-to-text transcription and dialect identification, focused on Egyptian dialect speech obtained from YouTube.

The MGB-1 English challenge training data covered the whole range of seven weeks of BBC TV output across four channels, resulting in about 1,600 hours of broadcast audio. In addition, several hundred million words of BBC subtitle text was provided for language modeling. The MGB-2 Arabic challenge training data comprised a total of 1,200 hours released with lightly supervised transcriptions. Moreover, 110M words obtained from the Al Jazeera Arabic website \url{aljazeera.net} were released for language modeling.

While the previous MGB challenges used mainstream broadcast media (BBC in MGB-1,  Al Jazeera in MGB-2) the Arabic MGB-3 challenge used YouTube recordings for two reasons: it is a platform that enables dialectal recordings to be harvested easily, and it also allows the collection of videos across different genres. Thus the MGB-3 Arabic Challenge extends the diversity of the data compared to previous MGB Challenges. This results in a relatively high baseline word error rate (WER) for  MGB-3 Arabic, compared to MGB-2 Arabic (see Section \ref{sec:baselinel}).  We thus targeted the following aspects in MGB-3 Arabic:
\begin{itemize}[noitemsep]
\item Dealing with languages which do not have well-defined orthographic systems, Egyptian Arabic in particular.
\item  Multi-genre scenarios: seven different genres are included in MGB-3 Arabic.
\item  Low-resource scenarios: only 16 hours of in-domain data was provided, split into adaptation, development and testing data.
\end{itemize}

The MGB-3 Arabic data comprised 16 hours of Egyptian Arabic speech extracted from 80 YouTube videos 
distributed across  seven genres: comedy, cooking, family/kids, fashion, drama, sports, and science talks (TEDx)
\footnote{TEDx Talks includes prepared talks of up to 18 minutes duration;  the chosen TEDx talks are in Egyptian dialect:  {\url{https://www.youtube.com/user/TEDxTalks}}}

We assume that the MGB-3 data is not enough by itself to build robust speech recognition systems, but could be useful for adaptation, and for hyper-parameter tuning of models built using the MGB-2 data. Therefore, we reused the MGB-2 training data in this challenge, and considered the provided in-domain data as (supervised) adaptation data.

The first task of the MGB-3 Arabic Challenge was to build a standard speech-to-text transcription system and to provide time-stamped word recognition results. This year, participants were asked to run their systems on the MGB-2 test set, in order to report progress compared to the MGB-2 challenge~\cite{ali2016mgb}, as well as the MGB-3 Egyptian Arabic test set.

The second task was Arabic Dialect Identification (ADI).  In this task, participants were supplied with more than 50 hours of labeled data across the five major Arabic dialects: Egyptian (EGY), Levantine (LAV), Gulf (GLF), North African (NOR), and Modern Standard Arabic (MSA). Participants were encouraged to use the 10 hours per dialect to label more data from both the MGB2 and MGB3 data. Dialectal data and baseline code were shared on Github\footnote{\url{https://github.com/qcri/dialectID}}. Overall accuracy, precision, and recall were used as evaluation criteria across the five dialects. 

%% file: data.tex
\subsection{Data for Speech Recognition}

To build the MGB-3 corpus YouTube clips from various Egyptian channels were selected.  The various genres are shown in Table \ref{tab:3groups}. Across the seven different genres a total of 80 videos were selected.  

From each video the first 12 minutes were selected.  Manually-identified non-speech segments were removed.  The resulting clips  were then distributed into adaptation, development, and testing groups, with the test set being a little larger then the other two sets.  Details can be seen in Table \ref{tab:3groups}.  The table also summarizes how much of the overall data contains overlapping speech (more than one speaker talking simultaneously) and how much data contains non-overlapping speech.

\begin{table}[t]
\begin{center}
\scalebox{0.9}{
\begin{tabular}{c|cccc}
 \multicolumn{1}{c}{ } & Adapt & Dev & Test  \\
 \hline
 Comedy & 0.6/3 & 0.6/3 & 1.0/5  \\
 Cooking & 0.6/3 & 0.8/4 &  0.6/3 \\
 FamilyKids & 0.8/4 & 0.6/3 & 1.0/5  \\
 Fashion & 0.6/3 & 0.6/3 &  0.8/4 \\
 Drama  & 0.6/3 & 0.8/4 & 0.8/4  \\
 Science & 0.6/3 & 0.8/4 & 1.0/5  \\
 Sports  & 0.8/4 & 0.6/3 & 0.8/4  \\
 \hline
 Total overlap speech segments* & 0.6& 0.3 & 0.5  \\
 Total non-overlap speech segments* & 4.0& 4.1& 5.3 \\
 \hline
 \textbf{Overall data}	 & \textbf{4.6/23}& \textbf{4.8/24} & \textbf{6.0/30}  \\
\hline
\end{tabular}
}
 \caption{MGB3 data distribution across the three classes, duration in hours/number of programs (12 minutes each). * is duration in hours across all speech segments}
 \label{tab:3groups}
\end{center}
\end{table}




It can be argued that Egyptian Arabic is a language with no specific orthographic rules \cite{aliwerd2017}.   Given that dialectal Arabic does not have a clearly defined orthography, different people tend to write the same word in slightly different forms. Therefore, instead of developing strict guidelines to ensure a standardized orthography, we allow for variations in spelling. Thus we decided to have multiple transcriptions, allowing transcribers to write the transcripts as they deemed correct.  This can be addressed in evaluation by using a multi-reference WER estimation (MR-WER)~\cite{ali2015multi}.

Table \ref{tab:interanno} shows the inter-annotator  disagreement on the development data. This table shows two numbers: the raw Word Error Rate (WER) and the WER after applying surface normalization\footnote{Surface orthographic normalization for three characters; alef, yah and hah, which are often mistakenly written in dialectal text. This normalization is standard for dialectal Arabic pre-processing and reduces the sparseness in the text.}. 
This indicates that there is about 13\% disagreement between the annotators for the MGB-3 data. We will report results for the MGB-3 data after normalization only.

\begin{table}[t]
\begin{center}
\begin{tabular}{c|cccc}
 \multicolumn{1}{c}{ } & ref2 & ref3 & ref4  \\
 \hline
 ref1 & 23/17 & 17/14 & 15/11  \\
 ref2 & -- & 19/15 &  20/16 \\
 ref3  & -- & -- & 8/7  \\
 \hline
\end{tabular}
 \caption{word-level inter annotator disagreement on the development data across the four different human references before/after normalization (in \%).}
 \label{tab:interanno}
\end{center}
\end{table}

\subsection{Data for Dialect Identification}
The Arabic Dialect Identification (ADI) task concerns the discrimination of speech at the utterance level between five Arabic dialects.

The MGB-3 Arabic ADI task was motivated by the success of the VarDial challenge \cite{malmasi2016discriminating,zampieri2017findings}, as well as the growing interest in dialectal Arabic in general. In VarDial 2016, participants were provided with input speech recognition transcripts, from which we further extracted and provided lexical features. In VarDial 2017, we enriched the task by further providing acoustic features and releasing the audio files. In the MGB-3 Arabic Challenge, we used both acoustic and lexical features.  In addition, we also provided the  acoustic features for the 1,200 hours of training data used in the MGB-2 Challenge. 

The speech transcription was generated by a multi-dialect LVCSR system trained on the MGB2 data~\cite{khurana2016qcri}. For the acoustic features, we released a 400-dimensional i-vector for each utterance. We extracted these i-vectors using bottleneck features (BNF) trained on 60 hours of speech data~\cite{ahmed2016automatic}. The data for the ADI task comes from a multi-dialectal speech corpus created from high-quality broadcast, debate and discussion programs from Al Jazeera, and as such contains a combination of spontaneous and scripted speech~\cite{wray2015crowdsource}. We collected the training data set from the Broadcast News domain. The audio recordings were carried out at 16 kHz. The recordings were then segmented to label overlap speech sections, also removing any non-speech parts such as music and background noise.  Although the test and the development data sets came from the same broadcast domain, the recording setup for the test and development data differed from that of the training data. We downloaded the test and the development data directly from the high-quality video server for Al Jazeera (brightcove) over the period  July 2014 -- January 2015, as part of QCRI{\textquotesingle}s Advanced Transcription Service (QATS) \cite{ali2014qcri}. Table \ref{tab:ADI_data} shows some statistics about the ADI training, development and testing data sets.
\begin{table}[t]
\center
\scalebox{0.65}{
\begin{tabular}{c|rrr|rrr|rrr}
& \multicolumn{3}{c|}{\bf Training} & \multicolumn{3}{c|}{\bf Development} & \multicolumn{3}{c}{\bf Testing}\\
\bf Dialect & \bf Ex. & \bf Dur. & \bf Words & \bf Ex. & \bf Dur. & \bf Words & \bf Ex. & \bf Dur. & \bf Words \\
\hline
EGY&3,093&12.4&76&298&2&11&302&2.0&12\\
GLF&2,744&10.0&56&264&2&12&250&2.1&12\\
LAV&2,851&10.3&53&330&2&10&334&2.0&11\\
MSA&2183&10.4&69&281&2&13&262&1.9&13\\
NOR&2,954&10.5&38&351&2&10&344&2.1&10\\
\hline
\bf Total &\bf 13,825&\bf 53.6&\bf 292&\bf 1,524&\bf 10&\bf 56&\bf 1,492&\bf 10.1&\bf 58\\
\hline
\end{tabular}
}
\caption{The ADI data: examples (Ex.) in utterances, duration (Dur.) in hours, and words in 1000s.}
\label{tab:ADI_data}
\end{table}

\begin{figure}[t]
\centerline{\includegraphics[width=8.0cm]{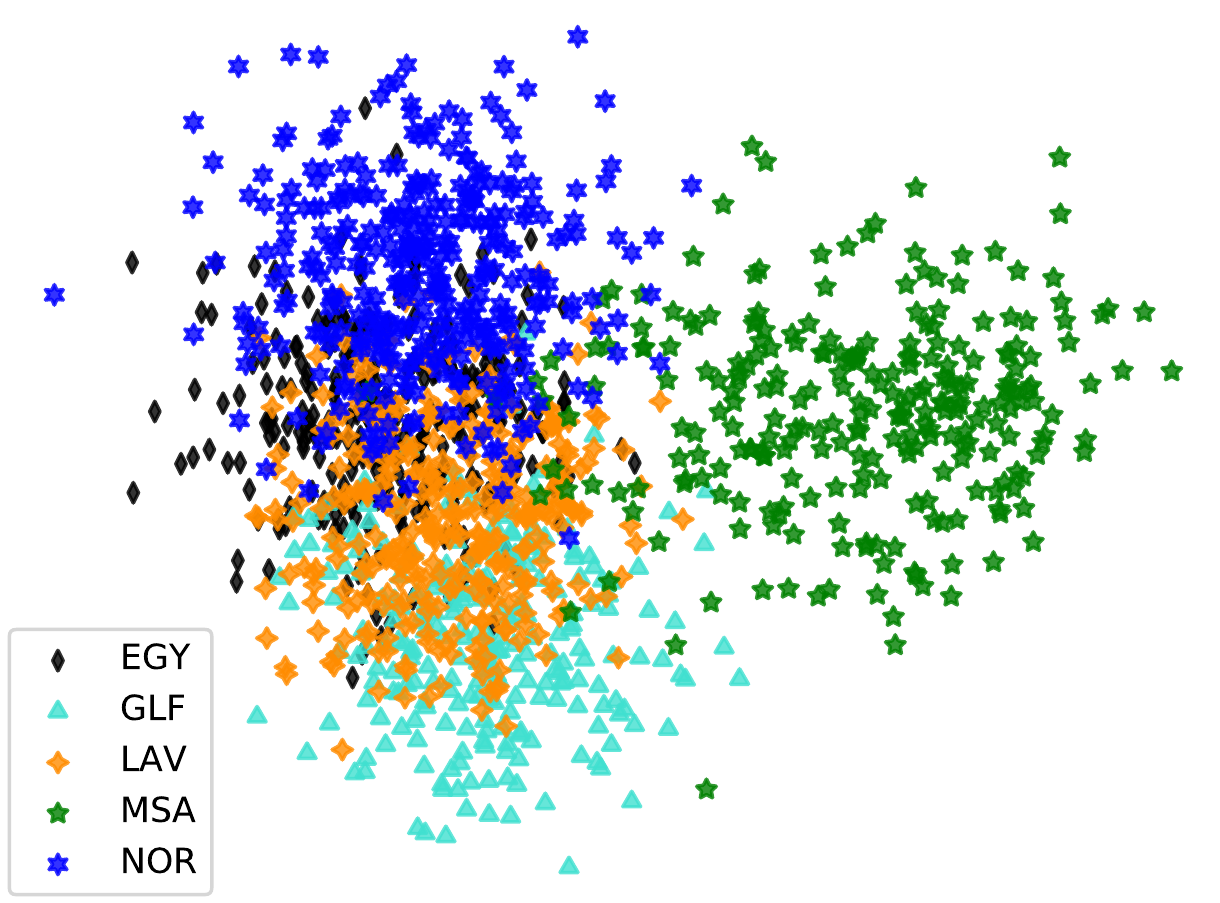}}
\caption{LDA projection for the 400 BNF i-vector feature vector for test data}
\label{fig:lda_test}
\end{figure}

Figure \ref{fig:lda_test} shows a  two dimensional Linear Discriminant Analysis (LDA) projection of the shared 400 dimension BNF i-vector for the test set, showing a clear separation between MSA, and the other four dialects. Both development and testing data have similar patterns in the same acoustics space.

%% file: baseline.tex
\label{sec:baselinel}
Similar to MGB-2, we provided an open source baseline systems for the challenge for both the speech transcription and dialect identification tasks. Word Error Rate (WER) continues to be the most commonly used metric for evaluating ASR. Usually, transcription of the speech signal is deterministic and one manual transcription should be a sufficient reference. However, as shown in table \ref{tab:interanno} there is about 13\% inter-annotator disagreement across the four annotators, which is to be expected as dialectal Arabic lacks a standardized orthography. For the MGB-3 Challenge, we investigated using MR-WER. This metric is based on comparing the recognized text against multiple manual transcriptions of the speech signal, which are all considered valid references. This approach thus accepts a recognized word if any of the references include it in the same form. The code for computing the MR-WER is available on GitHub\footnote{MR-WER \url{https://github.com/qcri/multiRefWER}.}.

\subsection{ASR Baseline}
The ASR baseline system was trained using the full MGB-2 data, 1,200 hours of audio.  This data was augmented by applying speed and volume perturbation \cite{ko2015audio}, increasing the number of training frames by a factor of 3. 
The code recipe is available on the Kaldi repository\footnote{\url{https://github.com/kaldi-asr/kaldi/tree/master/egs/gale_arabic/s5b}}. The acoustic modeling is similar to the QCRI submission to the MGB-2 Challenge \cite{khurana2016qcri}. The lexicon was grapheme-based, covering 950\,000 words collected from a set of shared lexicons, as well as the training data text.  The systems used a single-pass decoding with a trigram Language Model (LM), along with a purely sequence trained Time Delay Neural Network (TDNN) acoustic model~\cite{povey2016purely}; i-vectors were used for speaker adaptation.  We report results for the MGB-2 development set (5002 non-overlapping speech segments) on which we achieve a WER of 22.6\% without LM rescoring. This is a strong MGB-2 baseline compared to last year{\textquotesingle}s results. We also report results for the MGB-3 development set explained in table \ref{tab:3groups} using the MGB-2 baseline system, without adaptation to Egyptian Arabic using the MGB-3 data.

Table \ref{tab:ASRbaseline} shows the results for all the 1,279 non-overlapping speech  segments across the four annotators. We can observe that the MGB-3 baseline is poor, which is to be expected as the system was not adapted to the changed characteristics of the MGB-3 data.

\begin{table}[t]
\begin{center}
\scalebox{0.7}{
\begin{tabular}{c|ccccccc}
 \multicolumn{1}{c}{ } & WER1 & WER2 & WER3 & WER4 & AV-WER & MR-WER \\
 \hline
 Comedy & 59.5 & 59 & 58.8 &60.4& 59.4 &53.6 \\
 Cooking & 72.0 & 71.3 & 71.5 &71.2& 71.5&67.5\\
 FamilyKids & 50.2 & 48.4 & 48.3 &48.3& 48.8 &43.5 \\
 Fashion & 82.2 & 81.4 & 82.0 &81.2& 81.7 &78.0 \\
 Drama & 68.8&68.5&68.8&68.2& 68.6&64.5 \\
 Science & 59.3 & 57.7 & 59.5 &57.2& 58.4&51.4 \\
 Sports & 54.6 & 54.9 & 55.0 &54.4& 54.7& 49.4 \\
 \hline
 \bf Overall WER & \bf 63.8& \bf 62.9 & \bf 63.3 &\bf 62.8& \bf 63.20
& \bf 58.0\\
 \hline
\end{tabular}
}
 \caption{Baseline results in \% for the development data after applying surface text normalization }
 \label{tab:ASRbaseline}
\end{center}
\end{table}

\subsection{Dialect Identification Baseline}
The baseline for the ADI task uses a multi-class support vector machine (SVM) for classification. The lexical features were obtained from a speech recognition system, and bottleneck features were generated using the i-vector framework. The overall accuracy of this baseline system for the five-class ADI task is 57.2\%, with 60.8\% average precision and 58\% overall recall.

%% file: participant.tex
The MGB-3 Arabic Challenge sparked the interest of quite a number of research teams around the globe. The data was distributed to more than 20 teams, out of which
13 teams submitted 10 systems to the challenge for the ASR and ADI tasks. We attempt to highlight key features of the systems below. Detailed submitted system descriptions are cited and can be found on the MGB portal\footnote{\url{http://www.mgb-challenge.org/workshop.html}}. 

\subsection{Speech-to-Text Transcription}
In the ASR task, participants were asked to submit results for the MGB-2 and MGB-3 Arabic test sets. 
Participants submitted one primary submission and as many contrastive submissions as they wished. We scored and ranked results based on the primary submissions. The test set was manually segmented and only non-overlapping speech was used for scoring. 

\smallskip
\noindent
{\footnotesize\textbf{Aalto University} ({\url{spa.aalto.fi/en/research/research_groups/speech_recognition/}}): The novelties of the Aalto ASR system \cite{SmitMGB3} come from using TDNN-BLSTM acoustic models trained on 1,022 hours filtered from the MGB-2 training data, and adapted using the MGB-3 dialectal Egyptian data. Further improvement came from creating systems using subword and character-based language models (lexicon-free). The final submission was an minimum bayes risk (MBR)-decoded system combination of over 30 systems using two acoustic models and a variety of language models (character-, subword- and word-based). Aalto achieved the best results for both the MGB-2 (13.2\% WER) and scored 37.5\% AV-WER; 29.25\% MR-WER for MGB-3.

\smallskip
\noindent
\textbf{NDSC-THUEE:} 
The NDSC-THUEE system \cite{YangMGB3} used a TDNN followed by unidirectional LSTM layers or bidirectional LSTM (BLSTM) layers for the acoustic model. Their overall system makes use of speaker adaptive training, knowledge distillation-based domain adaptation, and MBR for system combination. Finally, they used an RNNLM for rescoring to generate their results. They achieved 14.5\% WER for MGB-2, and scored 40.8\% AV-WER and 32.5\% MR-WER for MGB-3.

\smallskip
\noindent
\textbf{Johns Hopkins University}
(JHU; {\url{clsp.jhu.edu}}): The JHU Kaldi system \cite{ManoharMGB3,povey2011kaldi} trained seed acoustic models using 982 hours filtered from the MGB-2 training set using speaker diarization and audio-transcript alignment, which was used to prepare lightly supervised transcriptions. They used a TDNN-LSTM acoustic model with a lattice-free (LF) MMI objective followed by segmental MBR (sMBR) discriminative training. For supervision, they fused transcripts from the four independent transcribers into confusion network training graphs. They achieved 16.0\% WER for MGB-2, and scored 40.7\% AV-WER and 32.8\% MR-WER for MGB-3.

\smallskip
\noindent
\textbf{MIT}
({\url{csail.mit.edu}}): The MIT system \cite{NajafianMGB3} used both the MGB-2 and MGB-3 data to train a wide range of acoustic models: DNN, TDNN,  LSTM, BLSTM, and prioritized grid LSTM (BPGLSTM) trained using using LF-MMI. They used both the Kaldi and CNTK toolkits. They applied 40 rounds of data augmentation to the MGB-3 data, and combined this with the MGB-2 data for acoustic domain adaptation. They used the full MGB-2 data without data filtering. They achieved 17.5\% WER for MGB-2, and scored 44.9\% AV-WER and 36.8\% MR-WER for MGB-3.

\smallskip
\noindent
\textbf{Brno University of Technology} 
(BUT; {\url{speech.fit.vutbr.cz}}): The BUT submission \cite{veselyMGB3} addressed the task as a low-resource challenge. Their system trained BLSTM-HMM models using 250 hours. They integrated speaker diariazation to improve speaker adaptation. They investigated the integration of the four transcriptions into acoustic model training, by using them serially (including each sentence four times into the training data, once with each transcription). An alternative, parallel, approach consisted of combining all the annotations into a confusion network. They achieved 24.0\% WER for  MGB-2, and scored 53.4\% AV-WER and 46.8\% MR-WER for MGB-3.

\smallskip
\noindent
\textbf{RDI \& Cairo University}
(RDI-CU; {\url{rdi-eg.com}}): The RDI-CU submission 
mainly focused on the MGB-2 task. The main specifications are the use of bottleneck (BN) features for training DNN-HMM models. For the acoustic model, they trained TDNN-BLSM, TDNN-LSTM and BLSTM, combining them using MBR. Both n-gram and  an RNN LM were used for rescoring. They achieved 16.0\% WER for MGB-2. Their MGB-3 performance was considerably worse compared to the MGB-2 results, with 62.7\% AV-WER  and 57.4\% MR-WER.
}

Table \ref{tab:ASR-systems} summarizes the main features of all the submitted systems. We can conclude that the leading teams benefited from transfer learning and audio adaption by building background acoustic models using the MGB-2 data and augmenting the five hours of in-domain MGB-3 training data. Also, language modelling approaches, such as lexicon adaption and  higher order n-gram and RNN LM rescoring, also made positive contributions to the overall systems. Only the Aalto team used subword language modeling to deal with the non-orthographic nature of the dialectal speech in the MGB-3 data. Finally, BUT and JHU explored combining the four transcriptions into a confusion matrix, allowing an alignment process to choose the best transcription.

Table \ref{tab:results_per_show} presents the error rates per genre for each of the submitted systems. In this table, we show both AV-WER across the four transcribers per genre, and the MR-WER. The most accurate system is consistently more accurate across all genres (with a small exception for the fashion genre). We also note that the ordering of systems by AV-WER and MR-WER can change, in particular at higher error rates. For example, results for comedy and science are not consistent between JHU and NDSC-THUEE. 
The overall ranking is still consistent using the two evaluation metrics. 
We ranked all the submitted systems with respect to MGB-2 WER (allowing direct comparison with the results of the 2016 MGB-2 Challenge), MGB3 AV-WER and MGB3 MR-WER in order. For MGB-2 the lowest WER reported in the challenge was reduced from 14.7\% in 2016 to 13.2\% on 2017.  Table \ref{tab:ASRFinalScore} summarizes the overall results, sorted by the best results. 

\begin{table*}[t]
\begin{center}
\begin{tabular}{c|ccccccc}
 \multicolumn{1}{c}{ } & Aalto & JHU & NDSC-THUEE & BUT & MIT & RDI-CU \\
 \hline
Used MGB2 data (in hours) & 1,022 & 982 & 680 & 250 & 1,200 & 500 \\
 MGB3 domain adaption (transfer learning) & \checkmark & \checkmark & \checkmark & \checkmark & \checkmark & - \\
Subword modeling & \checkmark & - & - & - & - & - \\
RNNLM rescoring & \checkmark & \checkmark & \checkmark& - & - & - \\
Speaker diarization & - & \checkmark & - & \checkmark & - & - \\
FST (confusion matrix) & - & \checkmark & - & \checkmark & - & - \\
Low-resource & - & - & - & \checkmark & - & - \\
\hdashline
AM (NN) &&&&&&  \\
LSTM & -& - & - & - & \checkmark & - \\
BLSTM & -& - & - & \checkmark & \checkmark & \checkmark  \\
BPGLSTM & -& - & - & - & \checkmark & - \\
TDNN & \checkmark & - & - & - & \checkmark & - \\
TDNN-LSTM & \checkmark & \checkmark & \checkmark & - & - & \checkmark  \\
TDNN-BLSTM & \checkmark & - & \checkmark & - & - & \checkmark  \\
\hline
\end{tabular}
 \caption{Main Features in the submitted systems for Arabic speech-to-text transcription.}
 \label{tab:ASR-systems}
\end{center}
\end{table*}

\begin{table*}[h!]
\begin{center}
\begin{tabular}{c|ccccccc}
 \multicolumn{1}{c}{ } & Aalto & JHU & NDSC-THUEE & BUT & MIT & RDI-CU \\
 \hline
 Comedy AV-WER & \textbf{51.4 }& 55.0 & 54.3 & 67.7 & 58.0 & 74.4 \\
 Comedy MR-WER & \textbf{42.4} & 45.7 & 46.2 & 61.5 & 50.0 & 69.4 \\
 \hline
 Cooking AV-WER & \textbf{38.2}& 43.1 & 43.8 & 57.1 & 46.7 & 72.8 \\
 Cooking MR-WER & \textbf{30.9 }& 36.1 & 37.2 & 52.0 & 40.1 & 69.9 \\
 \hline
 FamilyKids AV-WER & \textbf{30.6} & 35.3 & 33.9 & 49.6 & 38.0 & 61.92 \\ 
 FamilyKids MR-WER & \textbf{24.2} & 27.7 & 26.9 & 44.0 & 31.3 & 57.2 \\
 \hline
 Fashion AV-WER & 40.5 & 42.2 & \textbf{40.4} & 54.8 & 45.1 & 64.1 \\
 Fashion MR-WER & 30.9 & 31.5 & \textbf{30.9} & 46.9 & 35.44 & 57.1 \\
 \hline
 Drama AV-WER & \textbf{28.7} & 32.7 & 30.7 & 41.7 & 34.9 & 45.8 \\
 Drama MR-WER & \textbf{19.9} & 24.2 & 22.5 & 34.6 & 27.0 & 39.2 \\
 \hline
 Science AV-WER & \textbf{31.1} & 36.6 & 35.4 & 48.2 & 39.4 & 54.1 \\
 Science MR-WER & \textbf{23.1} & 27.7 & 27.2 & 41.4 & 31.6 & 47.6 \\
 \hline
 Sports AV-WER & \textbf{45.2} & 49.0 & 48.7 & 64.2 & 52.1 & 70.9 \\
 Sports MR-WER& \textbf{36.0} & 39.1 & 43.9 & 57.6 & 42.7 & 65.2 \\
 \hline
\textbf{MGB3 AV-WER}	 & \textbf{37.5}& \textbf{40.7} & \textbf{40.7} & \textbf{53.4}& \textbf{44.9} & \textbf{62.7} \\
\textbf{MGB3 MR-WER}	 & \textbf{29.3}& \textbf{32.8} & \textbf{32.5} & \textbf{46.8}&\textbf{ 36.8} &\textbf{ 57.7}\\
\hline
\end{tabular}
 \caption{Error rates (AV-WER and MR-WER over four reference transcriptions) per genre for Arabic speech-to-text transcription for the MGB-3 Egyptian Arabic test set.}
 \label{tab:results_per_show}
\end{center}
\end{table*}

\begin{table*}[h!]
\center
\begin{tabular}{l|c|cccccc}
 & \textbf{MGB2} & \multicolumn{4}{c}{\bf MGB3 WER per transcriber} & \multicolumn{2}{c}{\bf MGB3}\\
& \bf WER & \bf WER1 & \bf WER2 & \bf WER3 & \bf WER4 & \bf AV-WER & \bf MR-WER \\
\hline
\textit{2016-best system}&14.7&&&&&&\\ \cdashline{1-2}
\textbf{Aalto}&\textbf{13.2}& 38.0 & 37.7 & 37.4 & 36.9 & \textbf{37.5} & \textbf{29.3}\\
\textbf{NDSC-THUEE}&14.5& 41.5 & 40.1 & 40.7 & 40.8 & 40.75 & 32.5\\
\textbf{JHU}&16.0& 42.1 & 42.4 & 41.4 & 41.1& 40.7& 32.8 \\
\textbf{MIT}&17.5& 45.4 & 45.4 & 45.5 & 44.2 & 44.9 & 36.8\\
\textbf{BUT}&24.7& 55.0 & 55.2 & 54.3 & 54.4 & 53.4&46.8 \\

\textbf{RDI-CU}&16.0& 63.2 & 63.4 & 62.6 & 62.7 & 62.5 & 57.7\\
\hline
\end{tabular}
\caption{Summary of speech-to-text transcription results for MGB-2 and MGB-3 data.  For MGB-3, WERs are given for each of the four references (produced by different transcribers), as well as AV-WER and MR-WER across the four references.}
\label{tab:ASRFinalScore}
\end{table*}

\subsection{Arabic Dialect Identification}
In this task, participants were asked to label each sentence with one dialect. We received five submissions from eight teams. Similar to the ASR task, participants submitted one primary submission and as many contrastive submissions as they wished. We scored and ranked results based on the primary submissions. 

\smallskip
\noindent
{\footnotesize\textbf{MIT-QCRI} 
({\url{csail.mit.edu}; \url{qcri.org}}): The MIT-QCRI ADI system \cite{ShonMGB3} used both acoustic and linguistic features for the ADI challenge. They studied several approaches to address dialectal variability and domain mismatches between the training and testing data. They applied i-vector dimensionality reduction using a Siamese neural network. They also investigated using recursive whitening transformations. To optimize the network, they used a Euclidean distance loss function between the label and the cosine distance. For linguistic features, they used character features extracted from an LVCSR ASR system and phonetic features generated by phoneme recognizer; for the back-end they used an SVM. They achieved the best primary submission with 75.0\% overall accuracy, 75.1\% precision and 75.5\% recall.

\smallskip
\noindent
\textbf{University of Texas at Dallas} 
(UTD; \url{utdallas.edu}): The UTD submission \cite{BulutMGB3} was based on fusing five systems; four acoustic and one lexical. The first two systems used i-vector features extracted from MFCC features applied to two different classifiers, namely a Gaussian Back-end (GB) and a Generative Adversarial Network (GAN). The third system used the provided BNF i-vector features followed by the GB classifier. Their final acoustic system used unsupervised BNF (UBNF) for i-vector extraction followed by a GAN classifier. The fifth system explored lexical information by using unigram term frequency features followed by an SVM classifier. Their primary submission achieved 70.38\% overall accuracy, 71.7\% precision and 70.8\% recall. However, after releasing the reference test set, they conducted more experiments and achieved improved results, leading to the best results so far: an overall accuracy of 79.8\%, with 80.3\% precision and 79.9\% recall.

\smallskip
\noindent
\textbf{Singapore Institute of Technology \& K.J. Somaiya College, Mumbai}
(SG-MU): The SG-MU team 
developed three systems based on combinations of classifiers, namely a stacking classifier and two DNN based classifiers. The best results were obtained by the stacking classifier achieving 60.5\% overall accuracy, 61.6\% precision and 64.1\% recall by using just audio based i-vector features.

\smallskip
\noindent
\textbf{Birzeit University \& Fitchburg State University} : The BZ-FU team 
, they combined a number of different classifiers, each operating on one or the other of the shared data sources to achieve 69.3\% overall accuracy, 70.0\% precision and 69.5\% recall in the primary submission.

\smallskip
\noindent
\textbf{L'Universit\'e du Qu\'ebec \`a Montr\'eal} (UQAMl \url{uqam.ca}): In their primary submission, UQAM achieved 53.6\% overall accuracy, 54.3\% precision and 59.5\% recall.}

Table \ref{tab:ADIResults} summarizes the accuracy, precision, recall of the submitted systems for the ADI task. We  conclude that the leading systems deployed both acoustic and linguistic features to discriminate between different dialects. They also explored different techniques to reduce the impact of domain mismatch on the dialect classification task using GANs, Siamese networks, and recursive whitening transformations. 

\begin{table}[tb]
\begin{center}
\begin{tabular}{c|cccc}
 \multicolumn{1}{c}{ } & Acc & Prec & Rcl \\
 \hline
 MIT-QCRI & \textbf{75.0} & \textbf{75.1}& \textbf{75.5}\\
 UTD & 70.4 & 70.8 & 71.7 \\
 BZ-FU & 69.3 & 70 & 69.5  \\
 SG-MU & 60.5 & 61.6 & 64.1 \\
 UQAM & 53.6 & 54.3 & 59.5 \\
 \hdashline
 \textit{UTD*} & \textit{79.8} & \textit{79.9} & \textit{80.3} \\
 \hline
\end{tabular}
 \caption{ADI overall results (in \%) and ranking using primary submissions. \textit{UTD*} is a contrastive submission reported after releasing the test set, which is the best score achieved for this task.}
 \label{tab:ADIResults}
\end{center}
\end{table}